\documentclass[twoside,11pt]{article}

\usepackage{automl2019}

\usepackage{amssymb}
\usepackage{microtype}
\usepackage{graphicx}
\usepackage{booktabs} 
\usepackage{comment}
\usepackage{caption}

\usepackage{automl2019}

\jmlrheading{I. Drori, Y. Krishnamurthy, R. de Paula Lourenco, R. Rampin, K. Cho, C. Silva and J. Freire}

\ShortHeadings{AutoML by Pipeline Synthesis using Model-Based Reinforcement Learning and a Grammar}{Drori and Krishnamurthy and de Paula Lourenco and Rampin and Cho and Silva and Freire}
\firstpageno{1}

\begin{document}

\title{Automatic Machine Learning by Pipeline Synthesis using Model-Based Reinforcement Learning and a Grammar}

\author{\name{Iddo Drori} \email{idrori@nyu.edu} \\
       \name{Yamuna Krishnamurthy}\thanks{Y. Krishnamurthy is currently a doctoral candidate at Royal Holloway, University of London.\vspace{-10pt}} \email{yamuna@nyu.edu}\\
       \name{Raoni de Paula Lourenco} \email{raoni@nyu.edu}\\
       \name{Remi Rampin} \email{remi.rampin@nyu.edu}\\
       \name{Kyunghyun Cho} \email{kyunghyun.cho@nyu.edu}\\
       \name{Claudio Silva} \email{csilva@nyu.edu}\\
       \name{Juliana Freire} \email{juliana.freire@nyu.edu}\\
       \addr New York University }

\maketitle

\vspace{-10pt}

\begin{abstract}
Automatic machine learning is an important problem in the forefront of machine learning. The strongest AutoML systems are based on neural networks, evolutionary algorithms, and Bayesian optimization. Recently AlphaD3M reached state-of-the-art results with an order of magnitude speedup using reinforcement learning with self-play. In this work we extend AlphaD3M by using a pipeline grammar and a pre-trained model which generalizes from many different datasets and similar tasks. Our results demonstrate improved performance compared with our earlier work and existing methods on AutoML benchmark datasets for classification and regression tasks. In the spirit of reproducible research we make our data, models, and code publicly available.
\end{abstract}

\section{Introduction}
Machine learning has three main axes: dataset, task, and solution. Given a dataset, a well-defined machine learning task, and an evaluation criteria, the goal is to solve the task with respect to the dataset while optimizing performance. There are Automatic machine learning (AutoML) \citep{automl2019} problems of increasing difficulty, starting with hyperparameter optimization of a specific algorithm, to the selection of algorithms and their hyperparameter optimization, and finally meta learning, which entails synthesizing an entire machine learning pipeline from machine learning primitives. In this work we present a system that automates model discovery, solving well-defined machine learning tasks on unseen datasets by learning to generate machine learning pipelines from basic primitives. 

Our goal is to search within a large space of machine learning primitives and parameters, which together constitute a pipeline, for solving a task on a given dataset. The challenge is that the search space of pipelines, primitives and hyperparamaters is high dimensional. Similarly, programs for playing board games such as chess and Go are faced with the challenge of searching in a high dimensional space. We formulate the AutoML problem of pipeline synthesis as a single-player game, in which the player starts from an empty pipeline, and in each step is allowed to perform edit operations to add, remove, or replace pipeline components according to a pipeline grammar. A sequence of game steps results in a complete pipeline which is executed on the dataset to solve the task, evaluated by pipeline performance. Formally, an entire pipeline is a state, an action corresponds to modifying the current pipeline to derive a new pipeline, and pipeline performance is the reward. Thus, our approach is based on reinforcement learning \citep{sutton2018reinforcement}.

\begin{figure}[t]
\centering
\includegraphics[width=0.47\linewidth]{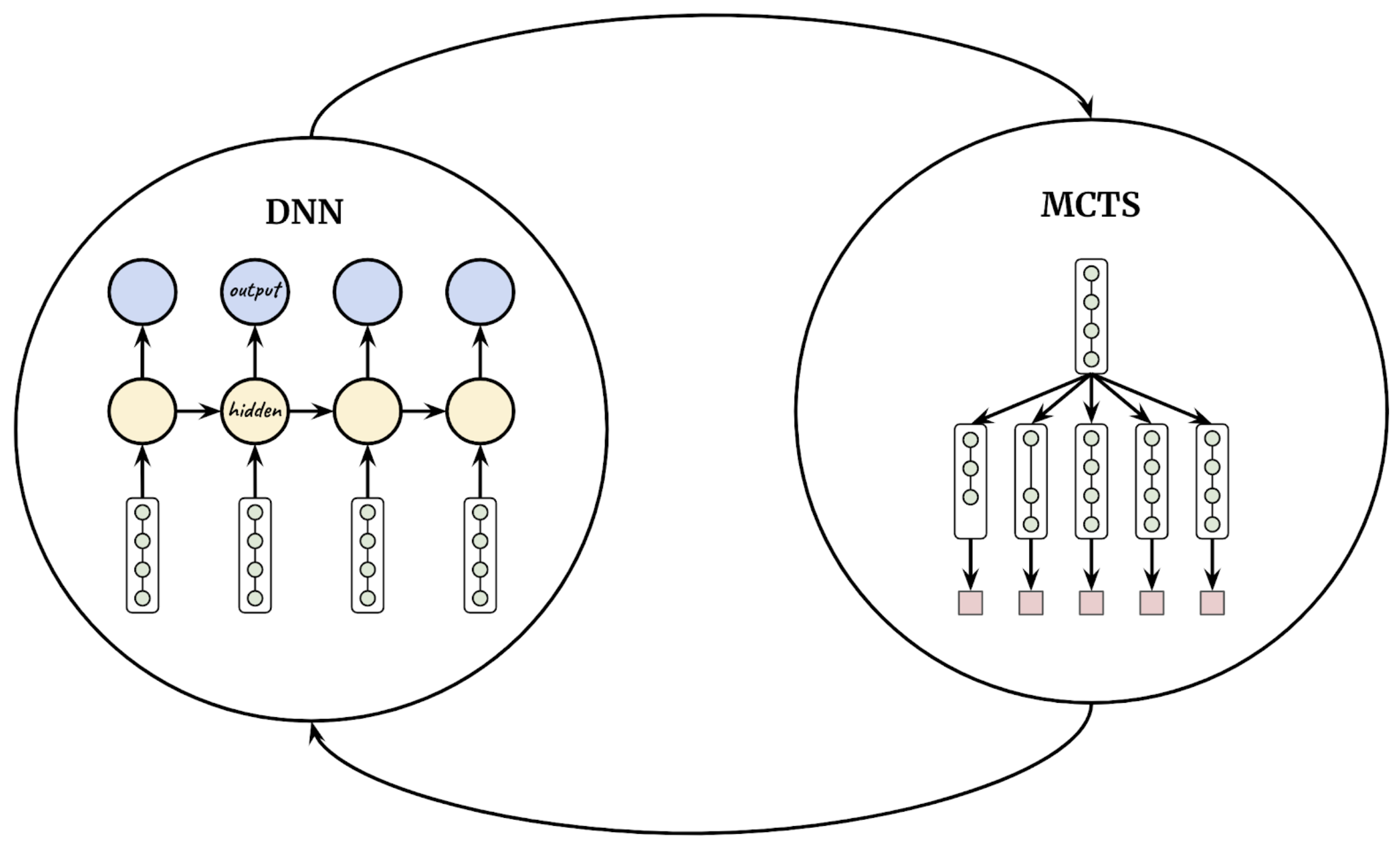}
\caption{Architecture: the neural network sequence model (left) receives an entire pipeline, meta features, and task as input. The network estimates action probabilities and pipeline evaluations. The MCTS (right) uses the network estimates to guide simulations which terminate at actual pipeline evaluations (square leaf nodes).}
\label{fig:overview}
\vspace{-15pt}
\end{figure}

An inherent advantage of this approach is that we have the provenance of all actions and decisions leading to a given pipeline, which provides an explanation for the synthesis process. Early programs for chess and Go searched a large space of millions of positions. AlphaZero \citep{silver2018general} reduced this search by three orders of magnitude using a general reinforcement learning algorithm. Inspired by expert iteration \citep{anthony2017thinking} we use a neural network for predicting pipeline performance and action probabilities along with a Monte-Carlo tree search (MCTS) which makes strong decisions based on the network, as shown in Figure \ref{fig:overview}. The process progresses by self play with iterative self improvement, and is known to be highly efficient at finding a solution to search problems in high dimensional spaces. Considering all combinations of machine learning primitives would result in many invalid pipelines. Therefore, we use a pipeline grammar which reduces the search space even further. Using a grammar decreases the branching factor and average depth of the MCTS. Our approach allows to learn to synthesize a pipeline from scratch, tabula rasa, or to generalize from many different datasets and similar tasks by using a pre-trained neural network. The main contributions of this work are:  
\begin{enumerate} 
\item \textbf{Pipeline grammar}: reducing the branching factor and average search depth. We compare performance across time, with and without a grammar.
\item \textbf{Pre-trained model}: generalizing from many different datasets and similar tasks in contrast with learning from scratch every time.
\item \textbf{Meta learning minute}: we investigate performance using benchmark AutoML datasets across time, focusing on the first minutes of computation. Our results demonstrate competitive performance with computation time which is an order of magnitude faster than the latest AutoSklearn.
\item \textbf{Open source}: in the spirit of reproducible research we make our data, models, and code publicly available \citep{codeAutoML2019}.
\end{enumerate}

Existing AutoML systems use methods such as differentiable programming \citep{mitar2017}, tree search \citep{atm2017}, evolutionary algorithm \citep{olson2016tpot, chen2018autostacker, gijsbers2018layered}, Bayesian optimization \citep{rasmussen2003gaussian, bergstra2012, feurer2015autosklearn, autoweka2017}, collaborative filtering \citep{yang2018oboe,fusi2018probabilistic}, and grammars \citep{de2017recipe}, for finding machine learning pipelines for a given dataset and task. Our system uses model-based reinforcement learning, combining a neural network with MCTS, and the search is constrained by a pipeline grammar. 

\section{Methods}
\label{sec:methods}
Our earlier work on AlphaD3M \citep{drori2018alphad3m}, formulates the AutoML problem of meta learning as pipeline synthesis using a single-player game with a neural network sequence model and Monte Carlo tree search (MCTS). A pipeline is a data mining work flow, of data pre-processing, feature extraction, feature selection, estimation, and post-processing primitives. The possible states are all valid pipelines generated from a set of primitives, constrained on actions of insertion, deletion or substitution of a primitive in the current pipeline. To reduce the search space, we define a pipeline grammar where the rules of the grammar constitute the actions. The grammar rules grow linearly with the number of primitives and hence address the issue of scalability. Our architecture models meta data and an entire pipeline chain as state rather than individual primitives. A pipeline, together with the meta data and problem definition is analogous to an entire game board configuration. The actions are transitions from one state (pipeline) to another, defined by the production rules of the grammar described next.

\subsection{Neural Network and Monte-Carlo Tree Search}
Our system uses a recurrent neural network shown in Figure \ref{fig:overview} (left), specifically a long short-term memory (LSTM) sequence model \citep{hochreiter1997long}, defined as $(p,v)=f_{\theta}(s)$ with parameters $\theta$. The network $f_{\theta}(s)$ receives a pipeline state representation as input $s$ and computes a vector of probabilities $p_{\theta}=P(a|s)$ over all valid actions $a$, and a value $v=\mathbb{E}[e|s]$ which approximates the pipeline evaluation $e$ when run on the data for solving the task. The system learns these action probabilities and the estimated values from games of self-play which guide the search in future games. The parameters $\theta$ are updated by stochastic gradient descent on the following loss function: 
\begin{equation}
\label{eq:lossfunction}
L(\theta) = - \pi \log p + (v - e)^{2} + \alpha \| \theta \|^{2},
\end{equation}
maximizing cross entropy between policy vector $p$ and search probabilities $\pi$, minimizing mean squared error between predicted performance $v$ and actual pipeline evaluation $e$, and regularizing the network parameters $\theta$ to avoid over fitting. Our system uses Monte-Carlo tree search which is a stochastic search using upper confidence bound update rule:
\begin{equation}
    U(s,a) = Q(s,a) + c P(a|s) \frac{\sqrt{N(s)}}{1 + N(s,a)},
\end{equation}
where $Q(s,a)$ is the expected reward for action $a$ from state $s$, $N(s,a)$ is the number of times action $a$ was taken from state $s$, $P(a|s)$ is the estimate of the neural network for the probability of taking action $a$ from state $s$, and $c$ is a constant which determines the amount of exploration. At each step of the simulation, we find the action $a$ and state $s$ which maximize $U(s,a)$ and add the new state to the tree, if it does not exist, with the neural network estimates $P(a|s),v(s)$ or call the search recursively. Finally, the search terminates and a pipeline is realized and applied to the data to solve the task, resulting in an actual evaluation $e$ of running the generated pipeline on the data and task.

\subsection{Pipeline Grammar}
\label{sec:grammar}
We use a context free grammar (CFG) \citep{chomsky1956three} to add domain knowledge for generating valid pipelines represented as strings. A CFG is represented by a four-tuple $<T, N, P, S>$ where $T$ is the set of terminals which are the components that make the string, $N$ the set of non-terminals which are placeholders for patterns of terminals that are generated from them, $P$ the set of production rules for replacing a non-terminal with other non-terminals or terminals and $S$ the start symbol, which is a special non-terminal symbol that appears in the initial string generated by the grammar. We use the prior knowledge of working pipelines for specific tasks in constructing the CFG for generating machine learning pipelines for that task. For example, for classification and regression tasks the pipeline usually consists of data cleaning, data transformation and an estimator (classifier or regressor) primitives in that order. Data cleaning consists of primitives that fix the data to make it suitable as inputs to the estimators, such as imputing missing values. Data transformation primitives transform the original data, for example, dimensionality reduction, categorical value encoding, and feature selection. Estimators are the learning primitives, which for a classification task are a set of classifiers such as Naive Bayes and SVM, and for regression are a set of regressors such as Linear and Ridge Regression. Based on this sequence of components for classification and regression tasks we define the CFG, using Sklearn \citep{pedregosa2011scikit} primitives, for classification as shown in Table \ref{tab:grammar}. Our system is configurable with different grammars for a given task.

\begin{table}
\centering
\footnotesize
\caption{Grammar $<T, N, P, S>$ for machine learning pipelines for a classification task.}
\label{tab:grammar}
\begin{tabular}{|l|l|}
\hline
$\textbf{T}$[Terminals] & $SkImputer$, $MissingIndicator$, $OneHotEncoder$, $OrdinalEncoder$, \\ 
& $PCA$ $\ldots$, $GaussianNB$, $RidgeClassifier$, $SGDClassifier$, $LinearSVC$ \\
\hline
$\textbf{N}$[Non-Terminals] & $DataCleaning$ ${<}DC{>}$, $DataTransformation$ ${<}DT{>}$, \\
& $Estimators$ ${<}E{>}$ \\
\hline
$\textbf{S}$[Start] & $S$ \\
\hline
$\textbf{P}$[Production Rules] & ${<}S{>}$\hspace{.4cm}::= ${<}E{>}$ $|$ ${<}DC{>}$ ${<}E{>}$ $|$ ${<}DT{>}$ ${<}E{>}$ $|$ ${<}DC{>}$ ${<}DT{>}$ ${<}E{>}$ \\
                      & ${<}DC{>}$\hspace{.1cm}::= $SkImputer$ ${<}DC{>}$ $|\ldots|$ $MissingIndicator$ ${<}DC{>}$ $|$ \\ 
                      & \hspace{1.7cm}$SkImputer$ $|\ldots|$ $MissingIndicator$ \\
                      & ${<}DT{>}$\hspace{.1cm}::= $OneHotEncoder$ ${<}DT{>}$  $OrdinalEncoder$ ${<}DT{>}$ $|\ldots|$ \\  & \hspace{1.7cm}$PCA$ ${<}DT{>}$ $|$\\
                      & \hspace{1.7cm}$OneHotEncoder$ $OrdinalEncoder$ $|\ldots|$ $PCA$\\
                      & ${<}E{>}$\hspace{.4cm}::= $GaussianNB$ $|$ $RidgeClassifier$ $|$ $SGDClassifier$ $|\ldots|$ \\
                      & \hspace{1.7cm} $LinearSVC$\\
\hline
\end{tabular}
\vspace{-15pt}
\end{table}

\section{Results}
\label{sec:results}
For evaluation we used 296 tabular datasets from OpenML \citep{openml2014} consisting of both classification and regression tasks. For our comparison with AutoSklearn as a baseline \citep{automlchallenges}, we used only comparable Sklearn primitives, specifically 2 data cleaning, 11 data transformation, 16 classification and 22 regression primitives. While AutoSklearn is limited to using the Sklearn library, AlphaD3M can be configured with machine learning primitives from other libraries.

\subsection{Pipeline Grammar}
\begin{figure}
\begin{minipage}[b]{0.5\linewidth}
\centering
\includegraphics[width=0.8\linewidth]{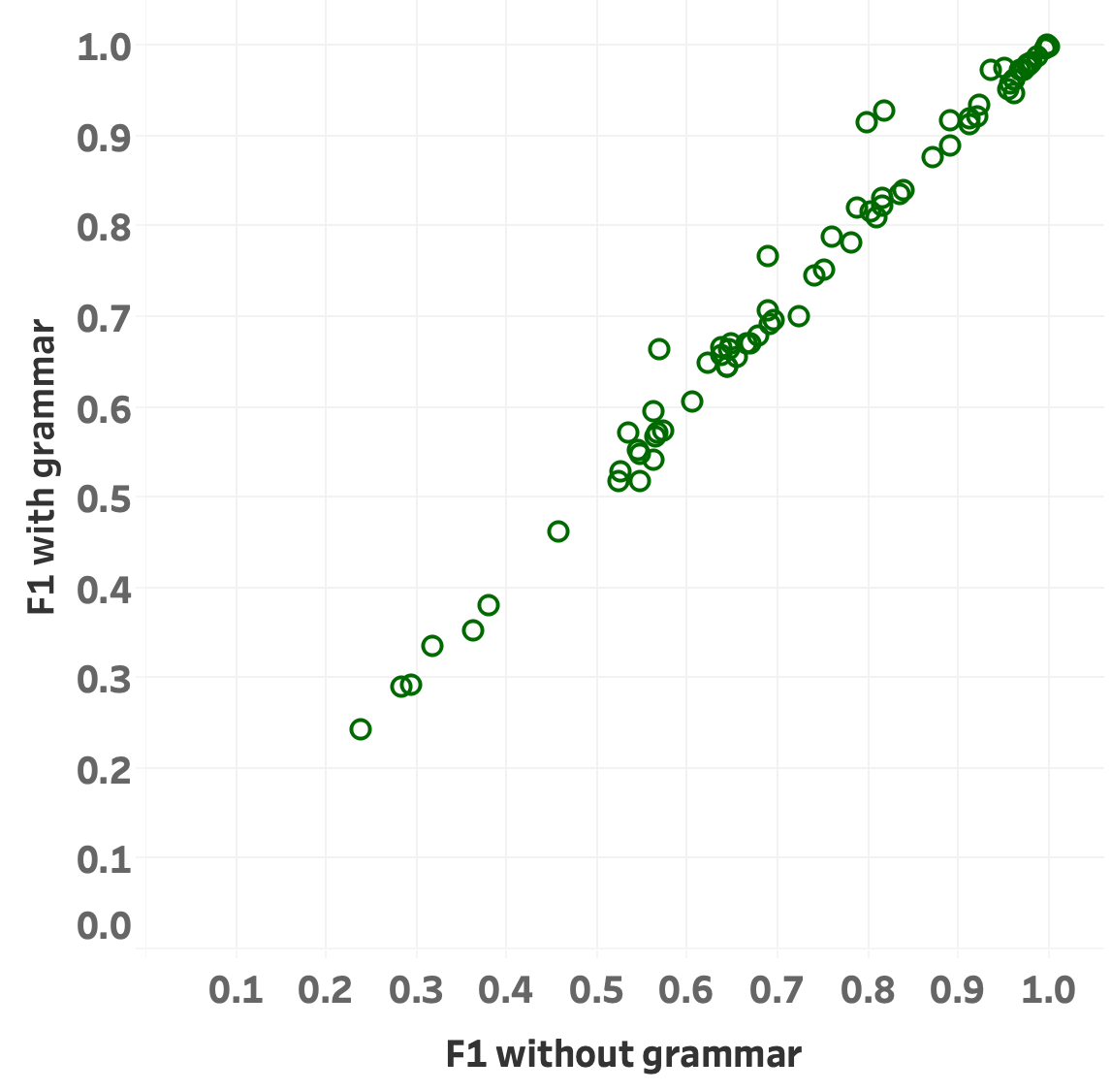}
\caption*{(a)}
\end{minipage}
\begin{minipage}[b]{0.5\linewidth}
\centering
\includegraphics[width=0.8\linewidth]{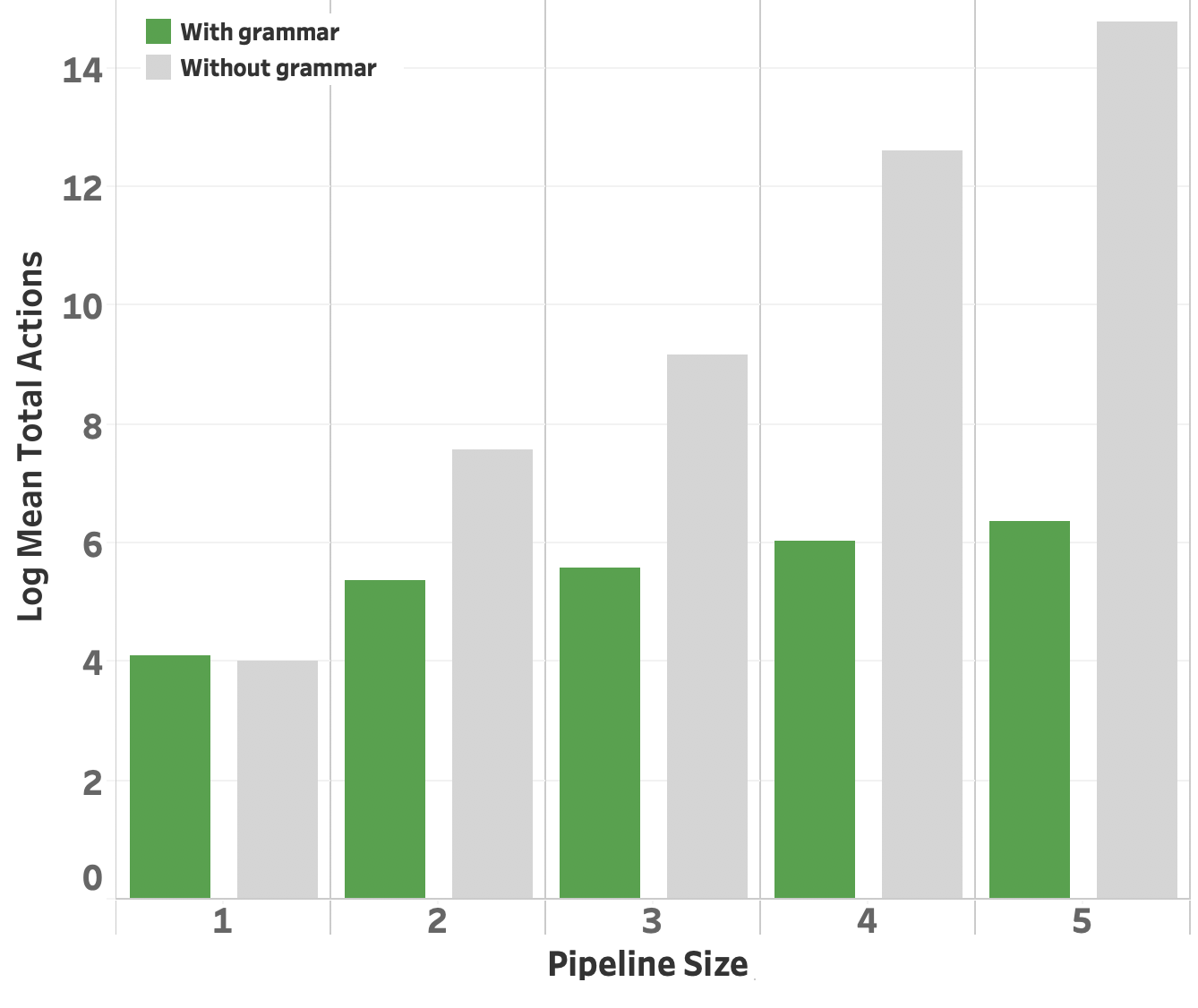}
\caption*{(b)}
\end{minipage}
\caption{(a) Comparison of performance with and without using a pipeline grammar: Each point represents an OpenML dataset. Performance is not degraded even though computation time is reduced. (b) Comparison of log mean total actions with and without a grammar.}
\label{fig:performancewwogrammar}
\vspace{-15pt}
\end{figure}

Figure \ref{fig:performancewwogrammar}(a) compares performance with and without a pipeline grammar. Each point represents one of 74 different OpenML datasets on a classification task, showing that performance is maintained, across the diagonal, even though using a grammar significantly reduces the search space. Figure \ref{fig:performancewwogrammar}(b) compares the logarithm of the mean of total actions with and without a pipeline grammar, showing a significant reduction in the search space size. Using the pipeline grammar is on average between two and three times as fast than without using the grammar. Using the grammar reduces the average depth of the MCTS by an order of magnitude and decreases the branching factor of the MCTS on average by three-fold compared to the non-grammar. 

\subsection{Pre-trained Model}
\label{sec:origresults}
\begin{figure}
\centering
\includegraphics[width=0.9\linewidth]{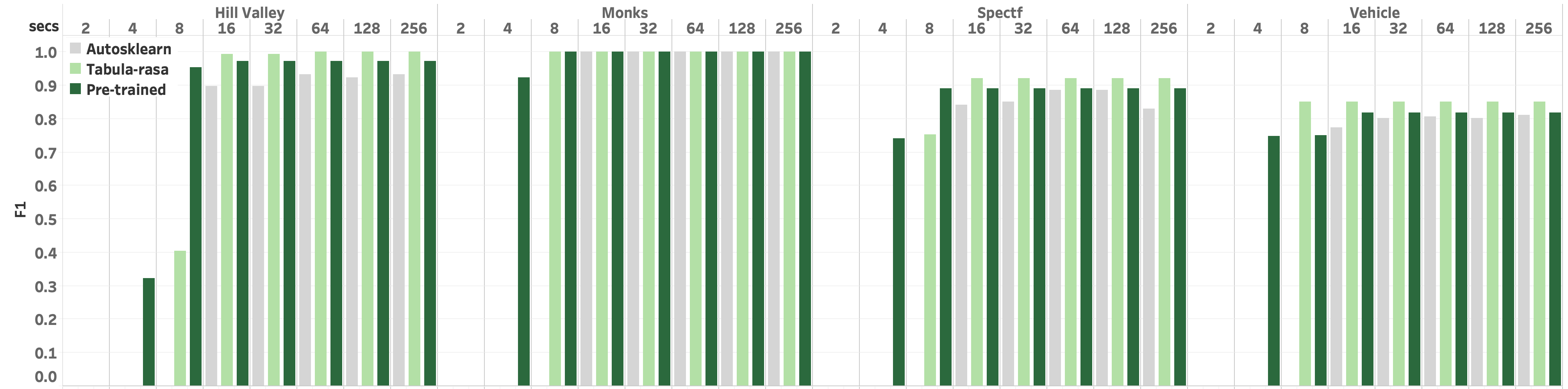}
\caption{Performance-time comparison between (i) AlphaD3M using a pre-trained model and a grammar, (ii) model trained tabula rasa and a grammar, (iii) AutoSklearn. All methods (including brute force) perform comparably given sufficient time using same primitives, the difference is in performance given equal times. Method (i) is faster than (ii) which in turn is faster than (iii). Performance is F1 and time is in seconds on an exponential scale.}%
\label{fig:classification-comparison}
\end{figure}

\begin{figure}
\centering
\includegraphics[width=0.4\linewidth]{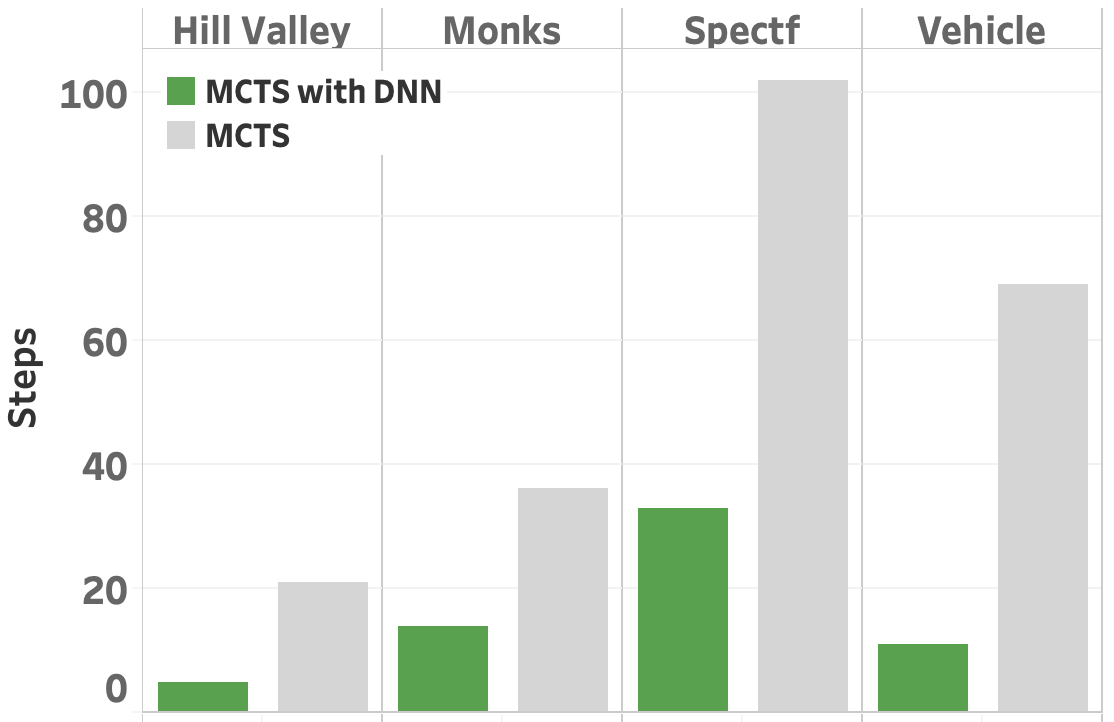}
\caption{Ablation comparison between number of steps of MCTS with NN vs. MCTS only.}%
\label{fig:mcts-comparison}
\vspace{-15pt}
\end{figure}

Comparing between AutoML methods requires taking into account both performance and running time. Given sufficient running time, most methods search the space until reaching an optimal solution, so our interest is in comparing between the efficiency of different methods. We therefore compare performance across multiple running times, which unveils how the methods progress. Specifically the most interesting difference occurs during the first minutes of computation. Figure \ref{fig:classification-comparison} compares performance between AlphaD3M using a grammar and a model pre-trained on other datasets (dark green), AlphaD3M using a grammar trained from scratch tabula rasa (light green), and AutoSklearn (gray). We used the same Sklearn machine learning primitives for both AlphaD3M and AutoSklearn, on a sample of benchmark AutoML datasets, running on a cluster with 4 Tesla 100 GPU's. The horizontal axis denotes time in $2^{i}$ seconds for $i=1,\ldots,8$, (ie left is better by an exponential factor in time). The vertical axis denotes F1-score in $[0,1]$, (ie higher is better). Our pre-trained model meta learns from other datasets and quickly generalizes reaching a result within 4 seconds (blue), including meta feature computation time. This is twice as fast as learning from scratch tabula rasa and our results demonstrate improvement in performance compared to learning the network from scratch. In turn, our method learning tabula rasa reaches a result twice as fast as AutoSklearn, with comparable performance. In AlphaD3M and AlphaZero, the neural network policy converges in the limit to the Monte-Carlo tree search (MCTS) policy, however the difference is in efficiency. Figure \ref{fig:mcts-comparison} compares the number of steps required for reaching the same performance using only MCTS with MCTS and a neural network.

\section{Conclusions}
\label{sec:conclusion}
We extended AlphaD3M, an automatic machine learning system by using a pipeline grammar and a pre-trained model. We analyzed the contribution of our extensions, comparing performance with and without a grammar, learning from scratch and using a pre-trained model, across time at fine granularity. Our results demonstrate competitive performance while being an order of magnitude faster than existing state-of-the-art AutoML methods. Our system is open, supporting any set of machine learning primitives, such as D3M, Sklearn, or general operations, thus opening the door for other application domains. In the spirit of reproducible research we make the data, models, and code available \citep{codeAutoML2019}.

\clearpage

\bibliography{bibliography}
\bibliographystyle{jmlr}
\end{document}